\documentclass[10pt,twocolumn,letterpaper]{article}

\usepackage{cvpr}              % To produce the CAMERA-READY version
\newcommand{\TODO}[1]{\textbf{\color{red}[TODO: #1]}}
\usepackage{multirow}
\renewcommand{\TODO}[1]{}

\usepackage{microtype}

\usepackage{colortbl}
\usepackage{arydshln,hhline}

\definecolor{cvprblue}{rgb}{0.21,0.49,0.74}
\usepackage[breaklinks,colorlinks,allcolors=cvprblue]{hyperref}

\def\paperID{31711} % *** Enter the Paper ID here
\def\confName{CVPR}
\def\confYear{2026}

\newcounter{tocmain}
\newcounter{tocsub}[tocmain]
\renewcommand{\thetocmain}{\Alph{tocmain}}
\renewcommand{\thetocsub}{\thetocmain.\arabic{tocsub}}
\newcommand{\TOCMain}[2]{%
  \refstepcounter{tocmain}%
  \setcounter{tocsub}{0}%
  \noindent\hyperref[#1]{\textcolor{black}{\textbf{\thetocmain. #2}}} \dotfill \hyperref[#1]{\textcolor{black}{p.\pageref{#1}}} \\
}
\newcommand{\TOCSub}[2]{%
  \refstepcounter{tocsub}%
  \noindent\hspace*{1.5em}\hyperref[#1]{\textcolor{black}{\thetocsub\ #2}} \dotfill \hyperref[#1]{\textcolor{black}{p.\pageref{#1}}} \\
}

\title{Detecting Unknown Objects via Energy-based Separation for Open World Object Detection}

\renewcommand{\thefootnote}{\fnsymbol{footnote}}
\author{Jun-Woo Heo\textsuperscript{1,*} \quad
Keonhee Park\textsuperscript{2,*,\dag} \quad
Gyeong-Moon Park\textsuperscript{1,\ddag}\\
$^{1}$Korea University, South Korea \quad $^{2}$Seoul National University, South Korea\\
{\tt\small dinleo11@korea.ac.kr, keonhee.park@vision.snu.ac.kr, gm-park@korea.ac.kr}
}
\renewcommand{\thefootnote}{\arabic{footnote}}

\begin{document}
\maketitle
\renewcommand{\thefootnote}{\fnsymbol{footnote}}
\footnotetext[1]{Equal contribution}
\footnotetext[2]{Work done at Korea University.}
\footnotetext[3]{Corresponding author}
\renewcommand{\thefootnote}{\arabic{footnote}}

% --------- Abstract ---------
\begin{abstract}
In this work, we tackle the problem of Open World Object Detection (OWOD). This challenging scenario requires the detector to incrementally learn to classify known objects without forgetting while identifying unknown objects without supervision.
Previous OWOD methods have enhanced the unknown discovery process and employed memory replay to mitigate catastrophic forgetting. 
However, since existing methods heavily rely on the detector's known class predictions for detecting unknown objects, they struggle to effectively learn and recognize unknown object representations.
Moreover, while memory replay mitigates forgetting of old classes, it often sacrifices the knowledge of newly learned classes. To resolve these limitations, we propose DEUS (\textbf{De}tecting \textbf{U}nknowns via energy-based \textbf{S}eparation), a novel framework that addresses the challenges of Open World Object Detection. DEUS consists of \textbf{E}quiangular Tight Frame (ETF)-Subspace \textbf{U}nknown \textbf{S}eparation (\textbf{EUS}) and an \textbf{E}nergy-based \textbf{K}nown \textbf{D}istinction (\textbf{EKD}) loss.
EUS leverages ETF-based geometric properties to create orthogonal subspaces, enabling cleaner separation between known and unknown object representations. Unlike prior energy-based approaches that consider only the known space, EUS utilizes energies from both spaces to better capture distinct patterns of unknown objects.
Furthermore, EKD loss enforces the separation between previous and current classifiers, thus minimizing knowledge interference between previous and newly learned classes during memory replay.
We thoroughly validate DEUS on OWOD benchmarks, demonstrating outstanding performance improvements in unknown detection while maintaining competitive known class performance.
\end{abstract}

% --------- Main Sections ---------
\vspace{-2mm}
\section{Introduction}
\label{sec:intro}
Object detection, a foundational task in computer vision, has achieved significant advances with the progress of deep learning~\cite{fang2021you,girshick2014rich,misra2021end,sun2021sparse}. 
However, traditional object detection approaches generally follow a closed-set paradigm, where the detector is restricted to recognizing only predefined classes during training. This closed-set setting hinders the detector from identifying objects that have not been encountered. To relax this restriction, Joseph \etal~\cite{joseph2021towards} introduced a new scenario, called Open World Object Detection (OWOD), in which the detector continuously learns annotated known objects while identifying unannotated objects as unknown. In this challenging scenario, when annotations for previously unknown objects become available, the detector must be incrementally updated to recognize unknown objects as known classes. Since supervision for unknown objects is not available in OWOD, the detector faces challenges in learning knowledge for unknown objects.

\begin{figure*}[t]
\centering
\begin{subfigure}[c]{0.28\textwidth}
    \centering
    \includegraphics[width=\textwidth, height=0.55\textheight, keepaspectratio]{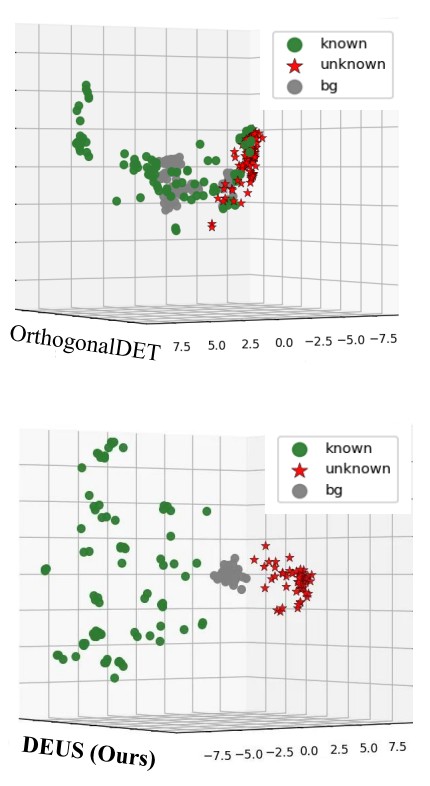}
    \caption{PCA of proposal features: baseline (top) vs. DEUS (bottom).}
    \label{fig:motiv_space}
\end{subfigure}
\hfill
\begin{subfigure}[c]{0.71\textwidth}
    \centering
    \includegraphics[width=\textwidth, keepaspectratio]{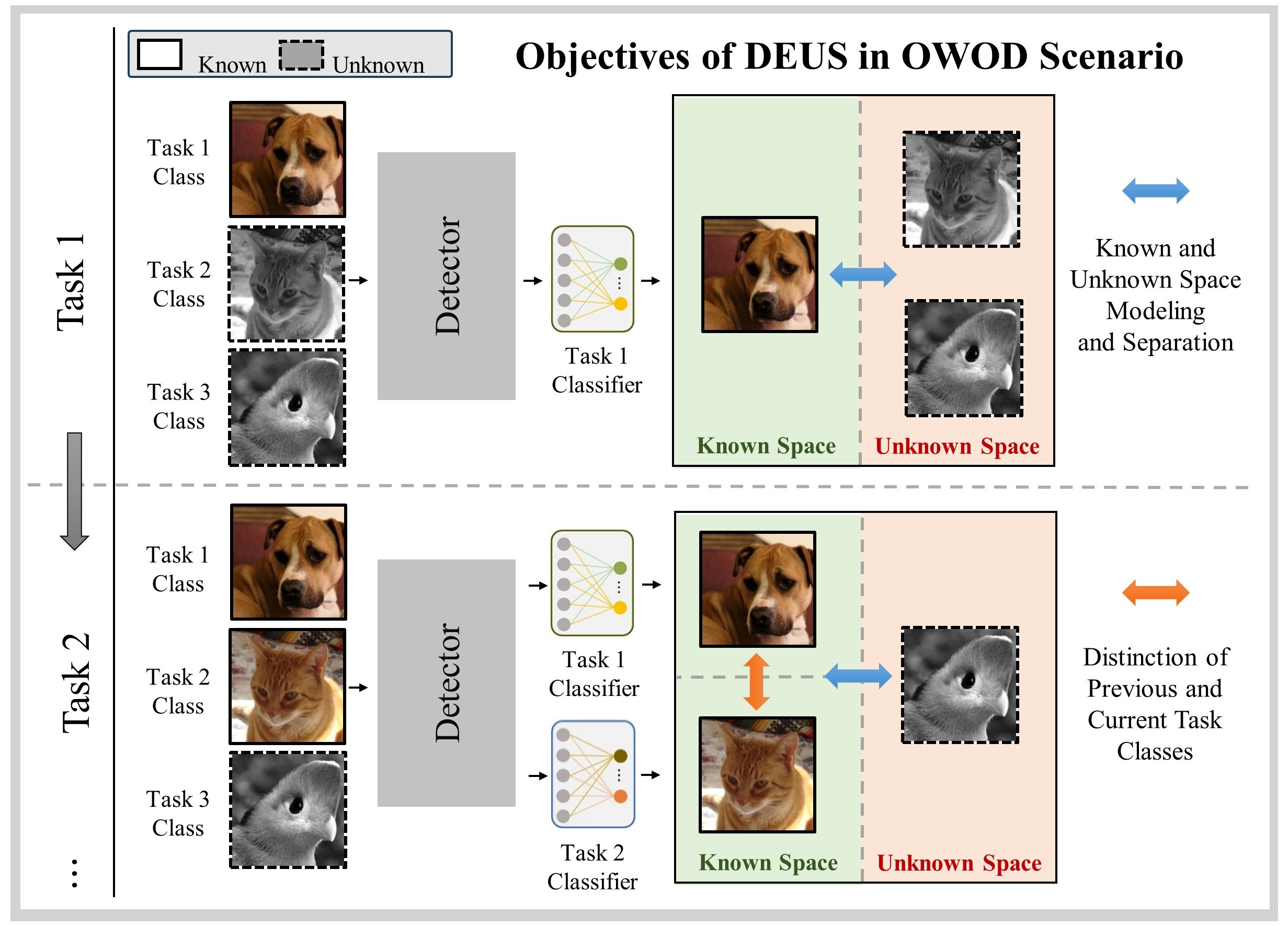}
    \caption{Objectives of DEUS in the OWOD scenario.}
    \label{fig:motiv_object}
\end{subfigure}
\caption{Motivation and objectives of DEUS. (a) PCA visualization of proposal features: the baseline (top) entangles known, unknown, and background, while DEUS (bottom) achieves clear separation. (b) Objectives of DEUS: separating known from unknown proposals while distinguishing previous and current known classes.}
\label{fig:motivation}
\end{figure*}
To address this, prior works~\cite{joseph2021towards,ma2023annealing,gupta2022ow,ma2023cat} propose an unknown discovery process, which utilizes the detector to assign pseudo-labels to specific regions in the background as unknowns. However, since this selection relies on the detector's current representations, it frequently selects partial areas of known objects or true background regions. This produces weak semantic pseudo-labels that blend known and unknown features, hindering effective discrimination. Since the model continuously learns from its own generated pseudo-labels during training, detecting higher-quality unknowns has become crucial in OWOD. To address this challenge, several approaches~\cite{liang2023unknown,du2022unknown, Zhang2025OpenWorldOM} integrate energy-based methods~\cite{liu2020energy} by using energy scores to better identify unknown regions. However, these methods also generally consider energy only within the known space, thus relying predominantly on known representations and the detector's known classification results. Such known space-only modeling approaches push away non-known objects from the known region but lack proper constraints to prevent unknown objects from being confused with background regions or vice versa, resulting in confused representations in the feature space as shown in the top of~\Cref{fig:motiv_space}. This leads to many unknown objects being overlooked or misclassified, preventing the learning of unknown representations.

Another issue in existing methods is that there is a trade-off between the performances of previous and current classes when learning new classes. In OWOD, while the detector learns novel classes sequentially for each task, it significantly forgets the previously learned classes, \textit{i.e.}, catastrophic forgetting arises. To address this, previous methods commonly adopt memory replay~\cite{mermillod2013stability, verwimp2021rehearsal, bonicelli2022effectiveness} to retain old knowledge while learning new classes. During memory replay, the detector is fine-tuned on both old and new classes to preserve old knowledge. Even though this replay effectively alleviates the forgetting issue of old classes, existing memory replay methods lack explicit regularization to prevent cross-influence between old and new classes during training. As growing task complexity increases the number of classes to be jointly optimized, this cross-influence becomes more severe, hindering effective learning of new classes while preserving knowledge of previously learned ones.

In this paper, we propose \textbf{DEUS}, a novel OWOD framework for \textbf{De}tecting \textbf{U}nknown objects via energy-based \textbf{S}eparation.
DEUS consists of \textbf{E}quiangular Tight Frame (ETF)-Subspace \textbf{U}nknown \textbf{S}eparation (\textbf{EUS}) and an \textbf{E}nergy-based \textbf{K}nown \textbf{D}istinction (\textbf{EKD}) loss. As illustrated in~\Cref{fig:motiv_object}, DEUS aims to address the two aforementioned challenges in the OWOD scenario by separating known and unknown proposals using EUS, and simultaneously distinguishing previous and current known classes with EKD.
First, EUS creates distinct known and unknown feature spaces to more effectively identify unknown objects. Unlike existing energy-based methods~\cite{liu2020energy,liang2023unknown,du2022unknown} that rely on a single known space (e.g., known classifier nodes), we jointly consider energies from two distinct Simplex ETF subspaces—one for the known space and one for the unknown space. During training, known and unknown proposals are each encouraged to attain high scores in their respective subspaces while scoring low in the opposite, and background proposals are guided toward the boundary region between the two subspaces. This bi-subspace energy learning guides features to naturally align with their respective spaces as shown in the bottom of~\Cref{fig:motiv_space} and enables the detector to capture discriminative knowledge, thereby effectively distinguishing each proposal.

Second, EKD is designed to alleviate the trade-off issue between the performances of old and new classes during memory replay. To compute energy scores separately, we partition known classifiers into old and new sub-classifiers. Here, higher energy scores indicate a stronger affinity to the corresponding classifier. For objects from old classes, the EKD loss encourages higher energy scores from the old sub-classifier and lower scores from the new one, and vice versa for new class objects. This energy-based constraint minimizes cross-influence between old and new classes during memory replay, enabling effective continual learning.
Through comprehensive experiments, we validate the effectiveness of DEUS, which achieves significantly improved unknown recall performance while balancing the learning of old and new classes during memory replay.

Our contributions can be summarized as follows:
\begin{itemize}[left=0.5em, itemsep=0pt]
    \item We propose \textbf{De}tecting \textbf{U}nknown objects via energy-based \textbf{S}eparation (DEUS), a novel OWOD framework that addresses two challenging issues in OWOD, limited unknown representation learning and cross-influence between old and new classes. 
    
    \item We introduce \textbf{E}TF-Subspace \textbf{U}nknown \textbf{S}eparation (EUS), the first approach to modeling geometrically separated distinct spaces and utilizing energy to separate known and unknown objects, helping to capture the knowledge of unknown objects and effectively discern unknowns from known or background.
    
    \item We design a new \textbf{E}nergy-based \textbf{K}nown \textbf{D}istinction (EKD) loss to alleviate the cross-influence between old and new classes during memory replay. This allows the detector to focus more on training each class set, enhancing the overall known performance.

    \item Experiments show that DEUS achieves state-of-the-art unknown recall across all benchmarks and tasks, while maintaining superior known mAP performance as the number of learned classes grows, demonstrating effectiveness in both unknown detection and continual learning.
    
\end{itemize}

\section{Related Work}
\label{sec:Related_work}
%-------------------------------------------------------------------------
\subsection{Open World Object Detection}
Joseph \etal~\cite{joseph2021towards} introduced Open World Object Detection (OWOD) to address the limitations of traditional closed-set object detection. OWOD faces challenges, as detectors often confuse known and unknown representation knowledge. Prior works have attempted to enhance unknown discovery by improving pseudo-labeling and objectness. Joseph \etal~\cite{joseph2021towards} used an RPN-based detector with an energy-based unknown identifier (EBUI), which required additional weak supervision of unknown objects. Gupta \etal~\cite{gupta2022ow} applied attention-driven matching for pseudo-labeling, while Ma \etal~\cite{ma2023annealing} proposed label transfer learning and annealing-based scheduling to separate known from unknown representation knowledge. Ma \etal~\cite{ma2023cat} decoupled localization and identification, introducing self-adaptive pseudo-labeling. Zohar \etal~\cite{zohar2023prob} adopted a normal distribution for class-agnostic objectness, and Sun \etal~\cite{sun2024exploring} reduced the correlation between objectness and class predictions via orthogonalization. However, due to the lack of supervision for unknown objects, prior works have not focused on learning representations specific to unknowns.
\subsection{Energy Score}
Energy-based methods~\cite{liu2020energy} have recently been widely adopted for out-of-distribution detection. The energy score, computed as the negative log-sum-exponential of logits, provides a unified measure for distinguishing in-distribution from out-of-distribution samples. Park \etal~\cite{park2025online} introduced an energy-guided discovery to identify novel categories within unlabeled data. In unknown object detection, Liang \etal~\cite{liang2023unknown} proposed a negative energy suppression loss to filter out non-object samples, while Du \etal~\cite{du2022unknown} and Zhang \etal~\cite{Zhang2025OpenWorldOM} introduced energy-based uncertainty regularization to model the uncertainty between known and unknown objects. In Open World Object Detection, Joseph \etal~\cite{joseph2021towards} proposed an energy-based classifier to distinguish known from unknown objects. However, existing energy-based approaches in OWOD primarily rely on known class predictions, lacking explicit modeling of unknown representations, which leads to confusion between unknown objects and background regions and misclassification of known object parts as unknowns.

\section{DEUS}
We propose \textbf{DEUS} (\textbf{De}tecting \textbf{U}nknown objects via energy-based \textbf{S}eparation), which effectively addresses the key challenges in Open World Object Detection. In Sec.~\ref{sec:3.1}, we first introduce the problem definition of Open World Object Detection. We then describe the pipeline of the base model in Sec.~\ref{sec:3.2} for better understanding. In Sec.~\ref{sec:3.3}, we propose the ETF-Subspace Unknown Separation (EUS) technique that models geometrically distinct known and unknown subspaces based on Equiangular Tight Frame (ETF)~\cite{papyan2020prevalence} and uses energy to guide objects to their respective spaces for effective separation. Finally, in Sec.~\ref{sec:3.4}, we introduce an Energy-based Known Distinction (EKD) loss to balance the learning of old and new classes during memory replay.

\begin{figure*}[th!]
  \centering
   \includegraphics[width=\linewidth, height=0.38\textheight, keepaspectratio]{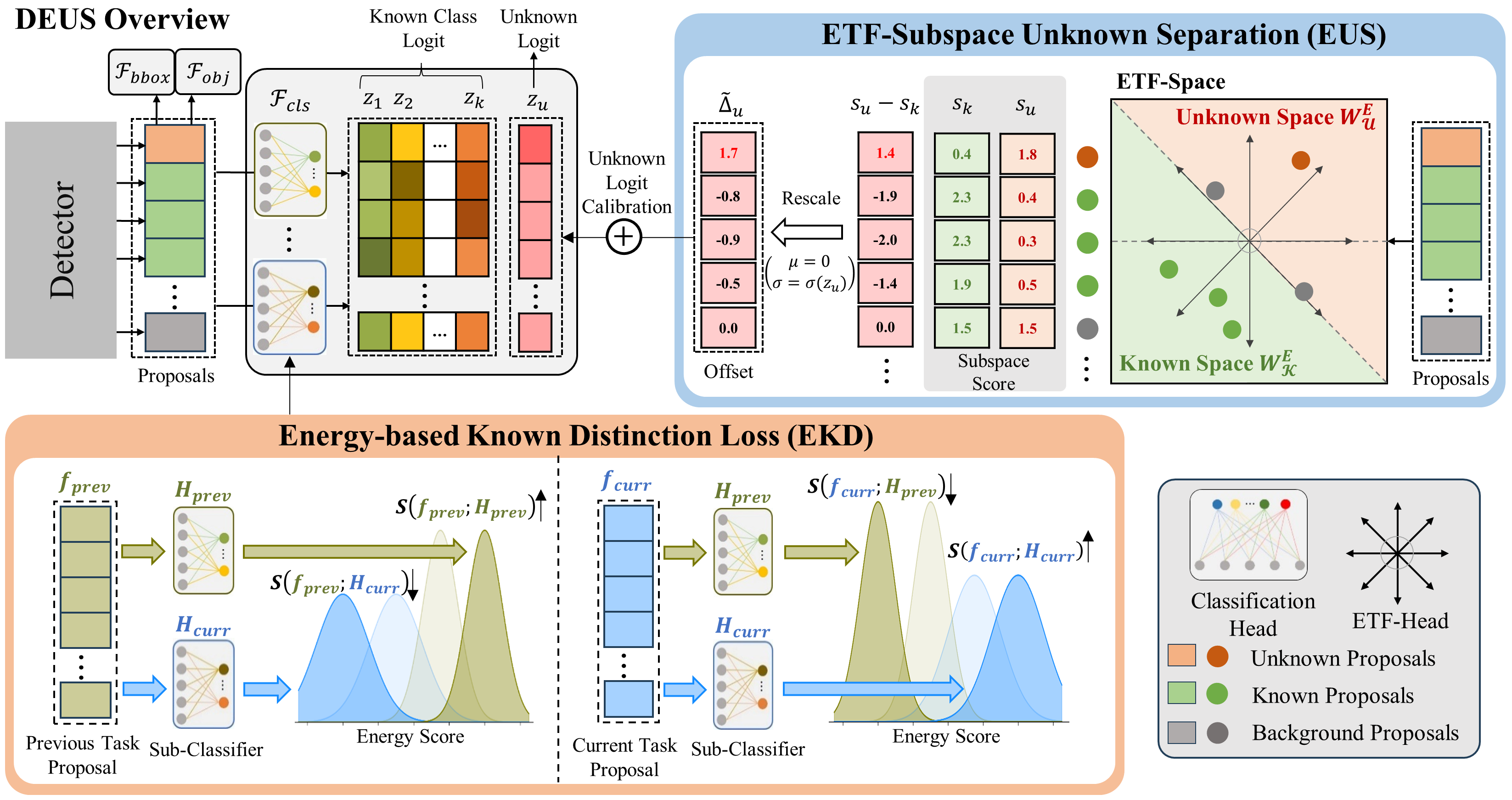}
   \caption{
   Overview of DEUS. ETF-Subspace Unknown Separation (EUS) utilizes a Simplex ETF to construct known and unknown spaces, where each space serves as an energy module to compute space energy scores, and Energy-based Known Distinction (EKD) loss is applied during the memory replay phase, where the classification branch is split into two sub-classifiers to calculate energy scores for previous and current tasks.
   }
   \label{fig:main}
\end{figure*}

\subsection{Problem Definition}
\label{sec:3.1}
In an Open World Object Detection (OWOD) task, a total of $T$ incremental tasks are sequentially given. In the $t$-th task, where $t \in \{1, \dots, T\}$, the detector is trained on dataset $\mathcal{D}_t = \{(\mathcal{I}_i^t, \mathcal{Y}_i^t)\}_{i=1}^N$ consisting of $N$ images, where $\mathcal{I}_i^t$ denotes the $i$-th input image and $\mathcal{Y}_i^t = \{c_j, b_j\}_{j=1}^{J_i}$ contains $J_i$ annotations for known objects. Here, $c_j$ denotes the class label, which belongs to the known class set $\mathcal{K}_t = \{1, 2, \dots, C\}$ where $C$ denotes the number of known classes at task $t$, and $b_j = [x_j, y_j, w_j, h_j]$ denotes the bounding box. The detector trained on task $t$ can identify known objects and detect objects from the unknown class set $\mathcal{U}_t = \{C + 1, \dots\}$ as unknowns. In the following task $t+1$, a subset of the unknown class set, $\mathcal{U}' = \{C + 1, \dots, C + n\}$, is labeled and merged into the updated known class set $\mathcal{K}_{t+1} = \mathcal{K}_t \cup \mathcal{U}'$, while the remaining unknown classes for task $t+1$ are given by $\mathcal{U}_{t+1} = \mathcal{U}_t \setminus \mathcal{U}'$. By repeating this process, the detector learns new classes incrementally, expanding its knowledge of known classes while continuously identifying unseen classes as unknowns.

\subsection{Pipeline of the Base Model}
\label{sec:3.2}
We adopt OrthogonalDet~\cite{sun2024exploring} as the base model due to its promising performance in identifying known objects in OWOD.
Given an input image, the image backbone extracts a feature map and the detector obtains object proposal features $f \in \mathbb{R}^d$ from the feature map using RoI pooling~\cite{girshickICCV15fastrcnn}, where $d$ is the feature dimension. The extracted proposal feature $f$ is fed into the bounding box~($\mathcal{F}_{bbox}$), objectness~($\mathcal{F}_{obj}$), and classification~($\mathcal{F}_{cls}$) branches to get:
\vspace{-2mm}
\begin{equation}
  z_{obj} = \mathcal{F}_{obj}(||f||),\; z_{cls} = \mathcal{F}_{cls}(\frac{f}{||f||}),\; z_{bbox} = \mathcal{F}_{bbox}(f),
\end{equation}
where $z_{obj} \in \mathbb{R}$ denotes the objectness score of the proposal $f$, and $z_{cls} \in \mathbb{R}^{C+1}$ denotes the logits for $C+1$ classes, which include $C$ known class nodes and a single unknown node, and $z_{bbox} \in \mathbb{R}^4$ denotes the bounding box regression outputs. Using these outputs, the base model forms joint class probabilities for each proposal as \(p_{jt}=\operatorname{softmax}(z_{cls}) \cdot \sigma(z_{obj})\), where $p_{jt}$ denotes joint probabilities, and $\sigma(\cdot)$ denotes the sigmoid function. During training, these joint probabilities are converted back to logits by \(z_{jt}=\log\!\frac{p_{jt}}{1-p_{jt}}\) for classification loss.
Given the ground-truth label as a one-hot vector $y_{gt}$, the classification loss is computed using the sigmoid focal loss~\cite{ross2017focal}:
\begin{equation}
\mathcal{L}_{\mathrm{cls}} = \mathcal{L}_{\mathrm{focal}}(y_{gt}, z_{jt}; \alpha, \gamma),
\label{eq:focal}
\end{equation}
where $\alpha$ and $\gamma$ are focal loss hyperparameters.
For the box branch, the regression loss is computed using $z_{bbox}$ with:
\begin{equation}
\mathcal{L}_{\mathrm{bbox}} = \mathcal{L}_{\mathrm{L1}}(z_{bbox}, b_{gt}) + \mathcal{L}_{\mathrm{gIoU}}(z_{bbox}, b_{gt}),
\end{equation}
where $b_{gt}$ denotes the ground-truth bounding box coordinates, $\mathcal{L}_{\mathrm{L1}}$ refers to the L1 loss, and $\mathcal{L}_{\mathrm{gIoU}}$ denotes the generalized IoU loss~\cite{rezatofighi2019generalized}.
Given the large number of proposals generated by the detector, it is essential to select appropriate proposals for training based on ground-truth. To achieve this, a dynamic matcher~\cite{wang2023random} is adopted to align object proposals with ground-truth, while a subset of the unmatched proposals is used as training targets for unknown objects.
\subsection{ETF-Subspace Unknown Separation}
\label{sec:3.3}
Existing OWOD methods have typically defined unknown objects based solely on the detector's known class predictions, without considering distinctive representations specific to unknown objects. This approach leads to missing unknown objects or misclassifying them as background regions. To address this, we propose the ETF-Subspace Unknown Separation (EUS), which explicitly builds two ETF-aligned subspaces (known / unknown) and guides each proposal toward its respective subspace.
To this end, we construct geometrically separated known and unknown spaces using a Simplex ETF, defined as:
\begin{equation}
 W^{E} \;=\; \sqrt{\frac{K}{K-1}}\;\Big( I_K - \frac{1}{K}\,\mathbf{1}_K \mathbf{1}_K^{\top} \Big)\, Q ,
\end{equation}
where $W^{E}\in\mathbb{R}^{K\times d}$ is the ETF basis matrix containing $K$ equiangular vectors, $\mathbf{1}_K$ is the all-ones vector, $Q \in \mathbb{R}^{K \times d}$ is an orthonormal matrix (i.e., $QQ^{\top} = I_K$), and $K$ is the number of ETF basis vectors. From the definition of Simplex ETF, we define known and unknown spaces as follows:
\begin{align}
&W^{E}_\mathcal{K} = [W^{E}_1;\dots;W^{E}_{\frac{K}{2}}] \in \mathbb{R}^{\frac{K}{2} \times d}, \qquad \\
&W^{E}_\mathcal{U} = [W^{E}_{\frac{K}{2}+1};\dots;W^{E}_{K}] \in \mathbb{R}^{\frac{K}{2} \times d},
\end{align}
where $W^{E}_\mathcal{K}$ and $W^{E}_\mathcal{U}$ denote non-overlapping subspaces from the Simplex ETF, and these bases are fixed and non-learnable.
Using the separated subspaces, the detector computes the energy with respect to the subspace. Each space functions as an energy module to compute the \textit{Helmholtz free energy}~\cite{liu2020energy} for a proposal feature $f$:
\vspace{-2mm}
\begin{align}
&E^\mathcal{K}(f) = -\log \sum_{i=1}^{K/2} \exp\big(W^{E}_{\mathcal{K},i} \cdot f\big), \qquad \\
&E^\mathcal{U}(f) = -\log \sum_{i=1}^{K/2} \exp\big(W^{E}_{\mathcal{U},i} \cdot f\big),
\label{ETF_Energy}
\end{align}
where $E^\mathcal{K}(f)$ and $E^\mathcal{U}(f)$ denote the known and unknown energies, respectively, while $W^{E}_{\mathcal{K},i} \cdot f$ and $W^{E}_{\mathcal{U},i} \cdot f$ represent the projections of the feature $f$ onto the $i$-th ETF basis vector in the known and unknown subspaces.
Then, we define subspace scores by negating energies:
\begin{equation}
s_{k}(f) = -E^\mathcal{K}(f),\qquad s_{u}(f) = -E^\mathcal{U}(f),
\label{score}
\end{equation}
where $s_k$ and $s_u$ denote the known and unknown scores indicating whether $f$ belongs to the known and unknown subspaces, respectively, where a higher value implies a stronger association with the corresponding subspace.
We then define the unknown offset as:
\begin{equation}
\Delta_u(f) = s_u(f) - s_k(f),
\label{delta}
\end{equation}
where $\Delta_u(f)$ represents how much higher the unknown subspace score is compared to the known subspace score for the feature $f$.
To guide known and unknown proposals toward their respective subspaces, EUS employs a loss function consisting of two complementary terms. First, we use an energy-based margin loss on the unknown offset $\Delta_u$:
\begin{equation}
\begin{split}
    &\mathcal{L}_{\mathrm{energy}} =
    \\
    &\mathbb{E}_{f}\left[
    \begin{cases}
    \max\left(0, m + \Delta_u(f)\right)^{2} & \text{if } f \text{ is GT-matched} \\
    \max\left(0, m - \Delta_u(f)\right)^{2} & \text{if } f \text{ is pseudo-unknown} \\
    0 & \text{otherwise}
    \end{cases}
    \right],
\end{split}
\end{equation}
where $m$ is a hyperparameter for the minimum margin gap. This loss enforces a margin between known and unknown scores by driving $\Delta_u(f) \le -m$ for known and $\Delta_u(f) \ge m$ for pseudo-unknown proposals.
For more stable convergence during training, we additionally adopt a focal loss (Eq.~\ref{eq:focal}) on the subspace scores $[s_k,s_u]$:
\begin{equation}
\mathcal{L}_{\mathrm{subspace}} \;=\; \mathcal{L}_{\mathrm{focal}}(t,[s_k,s_u] ;\alpha, \gamma),
\end{equation}
where the target one-hot vector is $t=[1,0]$ for GT-matched proposals, $t=[0,1]$ for pseudo-unknown proposals, and $t=[0,0]$ for background (remaining proposals unmatched to any ground-truth or pseudo-unknown).
The final EUS objective is the sum of the two terms:
\vspace{-1mm}
\begin{equation}
\mathcal{L}_{\mathrm{EUS}} \;=\; \mathcal{L}_{\mathrm{energy}} \;+\; \mathcal{L}_{\mathrm{subspace}} \, .
\end{equation}
While the energy loss provides the primary mechanism for known-unknown separation, the subspace loss guides known, unknown, and background proposals to their respective regions stably, encouraging background proposals toward the boundary between the two subspaces.
At inference time, following energy-based unknown scoring~\cite{liu2020energy}, we first calculate the unknown logit \(z_u = \log \sum_{c=1}^{C} \exp(z_c)\), where $z_c$ represents the logit for known class $c$ from the classification head $z_{cls}$. Next, to reflect the subspace scores, we calibrate $z_u$ with the unknown offset $\Delta_u(f)$. We form a calibration term that decreases the logit for known proposals and increases it for unknown proposals by standardizing $\Delta_u(f)$ per image across proposals, $\tilde{\Delta}_u(f) = (\Delta_u(f) - \mu_{\Delta_u})/\sigma_{\Delta_u}$, and then rescaling it by the standard deviation of $z_u$ (denoted as $\sigma_{z_u}$) to match the scale. Accordingly, the final unknown logit is given by
\begin{equation}
z_u' \;=\; z_u + \sigma_{z_u}\,\tilde{\Delta}_u(f) \, .
\label{Z_u}
\end{equation}
This calibration enhances the logit for unknown proposals while suppressing it for known proposals. For background proposals, which comprise most of the proposals, near-zero calibration values are obtained.
In contrast to previous energy-based approaches that depend solely on the detector's known class head prediction, EUS explicitly projects features onto two subspaces to effectively capture not only known features but also unknown representations, which prior works overlook.
\subsection{Energy-based Known Distinction Loss}
\label{sec:3.4}
In OWOD, memory replay is commonly used to mitigate catastrophic forgetting. Although this technique helps the detector to retain the knowledge of old classes, as the number of known classes grows and the classification problem becomes more complex, the cross-influence between old and new classes makes it difficult to preserve old knowledge and learn new concepts simultaneously.
To address this, we propose an \emph{Energy-based Known Distinction} (EKD) loss that reduces cross-influence during memory replay by explicitly separating the old and new classifiers and guiding each proposal to its corresponding classifier using energy scores. Concretely, we split the known class classifier into two sub-classifiers: the previous task sub-classifier $H_{\text{prev}}$ and the current task sub-classifier $H_{\text{curr}}$, each handling its own classes. For each proposal feature $f$, we define the negative energy-based score as follows:
\vspace{-4mm}
\begin{equation}
    S(f;H) \;=\; -E(f;H)
    \;=\; \log\!\sum_{c=1}^{C_H}\exp\!\big(z_c(f;H)\big),
\label{head_score}
\end{equation}
where $H\in\{H_{\text{prev}},H_{\text{curr}}\}$ denotes a sub-classifier, $z_c(f;H)$ is the logit for class $c$, and $C_H$ is the number of classes handled by sub-classifier $H$. Similar to the energy scores defined in Sec.~\ref{sec:3.3}, a larger $S$ indicates lower energy (i.e., stronger affinity) with the sub-classifier $H$, as higher logits in the log-sum-exponential formulation naturally correspond to stronger confidence within that classifier's domain. To minimize cross-interference, we encourage each sub-classifier to respond more strongly to its corresponding task samples. Specifically, let $f_{\text{prev}}$ and $f_{\text{curr}}$ denote proposals from previous and current tasks, respectively. We expect proposals from previous tasks to have higher scores with $H_{\text{prev}}$ than with $H_{\text{curr}}$, that is, $S(f_{\text{prev}};H_{\text{prev}}) > S(f_{\text{prev}};H_{\text{curr}})$, and vice versa for current task proposals. To this end, we design the EKD loss that enforces these preferences via pairwise loss as follows:
\begin{equation}
\begin{split}
    &\mathcal{L}_{\text{prev}} =
    \log\!\Big(1+\exp\!\big[S(f_{\text{prev}};H_{\text{curr}})-S(f_{\text{prev}};H_{\text{prev}})\big]\Big),\\
    &\mathcal{L}_{\text{curr}} =
    \log\!\Big(1+\exp\!\big[S(f_{\text{curr}};H_{\text{prev}})-S(f_{\text{curr}};H_{\text{curr}})\big]\Big),\\
    &\mathcal{L}_{\mathrm{EKD}} = \mathcal{L}_{\text{prev}} + \mathcal{L}_{\text{curr}} .
\end{split}
\end{equation}
This contrastive loss reduces the interference of $H_{\text{curr}}$ in previous task proposals and of $H_{\text{prev}}$ in current task proposals, effectively mitigating the cross-influence during memory replay.
The training objective combines all components:
\begin{equation}
\mathcal{L}_{\text{total}} = \mathcal{L}_{\text{cls}} + \mathcal{L}_{\text{bbox}} + \mathcal{L}_{\text{EUS}} + \mathcal{L}_{\text{EKD}},
\end{equation}
where $\mathcal{L}_{\text{EKD}}$ is applied only during memory replay phases when training on incremental tasks.
\newcolumntype{g}{>{\columncolor{Gray!15}}c}
\newcolumntype{b}{>{\columncolor{RoyalBlue!10}}c}
\newcolumntype{y}{>{\columncolor{yellow!15}}c}

\section{Experiments}

\begin{table*}[t]
\caption{Experimental results on M-OWODB (top) and S-OWODB (bottom). 
Results are reported in terms of mean average precision (mAP) for known classes, unknown class recall (U-Rec), and harmonic score (H-Score). 
The best performance is highlighted in bold, with the second-best performance underlined. $\dagger$ denotes reproduced results after correcting M-OWODB annotation duplication bug identified in~\cite{Yavuz_2024_ACCV}, which may differ from the originally reported numbers.}
\resizebox{1.\textwidth}{!}{%
\setlength{\tabcolsep}{2pt} % Default value: 6pt
\renewcommand{\arraystretch}{1.6} % Default value: 1
\begin{tabular}{l|cgy|cccgy|cccgy|ccc}
\specialrule{2pt}{0pt}{0pt}
Task IDs & \multicolumn{3}{c|}{Task 1} & \multicolumn{5}{c|}{Task 2} & \multicolumn{5}{c|}{Task 3} & \multicolumn{3}{c}{Task 4} \\
\specialrule{1pt}{0pt}{0pt}
 \large{Method} & \renewcommand{\arraystretch}{1.2}{\begin{tabular}[c]{@{}c@{}}Current\\mAP\end{tabular}} & U-Rec & H-Score & \renewcommand{\arraystretch}{1.2}{\begin{tabular}[c]{@{}c@{}}Previous\\mAP\end{tabular}} & \renewcommand{\arraystretch}{1.2}{\begin{tabular}[c]{@{}c@{}}Current\\mAP\end{tabular}} & \renewcommand{\arraystretch}{1.2}{\begin{tabular}[c]{@{}c@{}}Known\\mAP\end{tabular}} & U-Rec & H-Score & \renewcommand{\arraystretch}{1.2}{\begin{tabular}[c]{@{}c@{}}Previous\\mAP\end{tabular}} & \renewcommand{\arraystretch}{1.2}{\begin{tabular}[c]{@{}c@{}}Current\\mAP\end{tabular}} & \renewcommand{\arraystretch}{1.2}{\begin{tabular}[c]{@{}c@{}}Known\\mAP\end{tabular}}& U-Rec & H-Score & \renewcommand{\arraystretch}{1.2}{\begin{tabular}[c]{@{}c@{}}Previous\\mAP\end{tabular}} & \renewcommand{\arraystretch}{1.2}{\begin{tabular}[c]{@{}c@{}}Current\\mAP\end{tabular}} & \renewcommand{\arraystretch}{1.2}{\begin{tabular}[c]{@{}c@{}}Known\\mAP\end{tabular}}\\
\hline
ORE~\cite{joseph2021towards} & 56.0 & 4.9 & 9.0 & 52.7 & 26.0 & 39.4 & 2.9 & 5.4 & 38.2 & 12.7 & 29.7 & 3.9 & 6.9 & 29.6 & 12.4 & 25.3 \\
OW-DETR~\cite{gupta2022ow} & 59.2 & 7.5 & 13.3 & 53.6 & 33.5 & 42.9 & 6.2 & 10.8 & 38.3 & 15.8 & 30.8 & 5.7 & 9.6 & 31.4 & 17.1 & 27.8 \\
CAT~\cite{ma2023cat} & 60.0 & 23.7 & 34.0 & 55.5 & 32.7 & 44.1 & 19.1 & 26.7 & 42.8 & 18.7 & 34.8 & 24.4 & 28.7 & 34.4 & 16.6 & 29.9 \\
PROB$^{\dagger}$~\cite{zohar2023prob} & \textbf{66.4} & 28.3 & 39.7 & \textbf{62.6} & 39.2 & 50.9 & 26.4 & 34.8 & 49.6 & 33.5 & 44.2 & 29.3 & 35.2 & 44.0 & 26.5 & 39.7 \\
OrthogonalDet$^{\dagger}$~\cite{sun2024exploring} & 65.1 & 36.3 & 46.6 & 58.2 & 44.2 & 51.2 & 30.2 & 38.0 & \underline{50.9} & \underline{40.1} & \underline{47.3} & 28.7 & 35.7 & \underline{49.1} & \underline{31.5} & \underline{44.7} \\
O1O~\cite{Yavuz_2024_ACCV}& 65.1 & \underline{49.3} & \underline{56.1} & \underline{61.0} & \underline{45.0} & \underline{53.0} & \underline{50.3} & \underline{51.6} & 50.0 & 36.5 & 45.5 & \underline{49.5} & \underline{47.4} & 46.2 & 31.0 & 42.4 \\
OWOBJ~\cite{Zhang2025OpenWorldOM}& 61.4 & 23.6 & 34.1 & 58.4 & 34.4 & 45.7 & 23.8 & 31.3 & 44.8 & 27.8 & 38.8 & 25.1 & 30.5 & 36.4 & 20.7 & 32.0 \\
\hdashline
\textbf{DEUS (Ours)} & \underline{66.2} & \textbf{65.1} & \textbf{65.6} & \underline{61.0} & \textbf{45.7} & \textbf{53.3} & \textbf{66.2} & \textbf{59.0} & \textbf{53.4} & \textbf{43.3} & \textbf{50.1} & \textbf{69.0} & \textbf{58.0} & \textbf{50.5} & \textbf{32.8} & \textbf{46.0}\\
\hline \hline
ORE~\cite{joseph2021towards} & 61.4 & 1.5 & 2.9 & 56.5 & 26.1 & 40.6 & 3.9 & 7.1 & 38.7 & 23.7 & 33.7 & 3.6 & 6.5 & 33.6 & 26.3 & 31.8 \\
OW-DETR~\cite{gupta2022ow} & 71.5 & 5.7 & 10.6 & 62.8 & 27.5 & 43.8 & 6.2 & 10.9 & 45.2 & 24.9 & 38.5 & 6.9 & 11.7 & 38.2 & 28.1 & 33.1 \\
CAT~\cite{ma2023cat} & \underline{74.2} & 24.0 & 36.3 & \underline{67.6} & 35.5 & 50.7 & 23.0 & 31.6 & 51.2 & 32.6 & 45.0 & 24.6 & 31.8 & 45.4 & 35.1 & 42.8 \\
PROB~\cite{zohar2023prob} & 73.4 & 17.6 & 28.4 & 66.3 & 36.0 & 50.4 & 22.3 & 30.9 & 47.8 & 30.4 & 42.0 & 24.8 & 31.2 & 42.6 & 31.7 & 39.9 \\
OrthogonalDet~\cite{sun2024exploring} & 71.6 & 24.6 & 36.6 & 64.0 & 39.9 & 51.3 & 27.9 & 36.1 & \underline{52.1} & \underline{42.2} & \underline{48.8} & 31.9 & 38.6 & \underline{48.7} & 38.8 & \underline{46.2} \\
O1O~\cite{Yavuz_2024_ACCV} & 72.6 & \underline{49.8} & \underline{59.1} & 65.3 & \textbf{44.9} & \underline{54.6} & \underline{51.1} & \underline{52.8} & 49.5 & 41.5 & 46.8 & \underline{48.1} & \underline{47.4} & 47.3 & \underline{42.0} & 45.9 \\
OWOBJ~\cite{Zhang2025OpenWorldOM} & \textbf{76.2} & 22.3 & 34.5 & \textbf{69.8} & 41.0 & \textbf{54.8} & 28.7 & 37.7 & 50.6 & 35.7 & 46.8 & 30.9 & 37.2 & 46.7 & 36.9 & 43.2 \\
\hdashline
\textbf{DEUS (Ours)} & 71.6 & \textbf{68.7} & \textbf{70.1} & 63.5 & \underline{43.0} & 52.7 & \textbf{62.9} & \textbf{57.4} & \textbf{53.6} & \textbf{45.4} & \textbf{50.9} & \textbf{60.7} & \textbf{55.4} & \textbf{50.7} & \textbf{42.8} & \textbf{48.8} \\
\specialrule{2pt}{0pt}{0pt}
\end{tabular}%
}
\label{tab:main}
\end{table*}

\subsection{Experimental Settings}
\paragraph{Datasets and Metrics.} We evaluated DEUS on two standard OWOD benchmarks, M-OWODB~\cite{joseph2021towards} and S-OWODB~\cite{gupta2022ow}, each consisting of four non-overlapping incremental tasks. We additionally constructed RS-OWODB, a remote sensing benchmark using the DIOR~\cite{li2019object} dataset with 5 classes per task, to evaluate generalizability beyond natural images (see Appendix A for benchmark details). For known classes, we used mean average precision (mAP) as the metric, which measures mAP for previously learned classes, currently learned classes, and all known classes. For unknown classes, we used the recall (U-Rec) as the main metric to evaluate the performance of detecting unknown objects. To measure the overall performance, considering both known and unknown objects, we adopted the harmonic mean (H-Score), combining the mAP of known and the recall of unknown classes. 

\paragraph{Implementation Details.} We used OrthogonalDet~\cite{sun2024exploring} as our base model. The weights of the EUS and EKD losses are set to 1.0. We set $K$ for the Simplex ETF spaces to 128, constructing both known and unknown spaces with 64 vectors each. Our implementation is based on MM-Detection~\cite{mmdetection}. In OWOD, memory replay and pseudo-labeling of unknown objects are commonly adopted to mitigate catastrophic forgetting and provide supervision for unknowns, respectively. We follow the standard memory replay protocol, storing exemplar samples per class for subsequent replay phases. For pseudo-labeling, we introduce an improved process that dynamically scales the number of pseudo-labels based on the known class count and filters noisy detections, enabling more accurate unknown targets. For further details on both memory replay settings and pseudo-labeling, please refer to the supplementary materials (see Appendix A).

\subsection{Experimental Results}
We present the comparison results for M-OWODB (top) and S-OWODB (bottom) in \Cref{tab:main}, comparing our DEUS with previous OWOD methods \cite{joseph2021towards,gupta2022ow,ma2023cat,zohar2023prob,sun2024exploring,Zhang2025OpenWorldOM,Yavuz_2024_ACCV}.
In OWOD benchmarks, a fundamental trade-off exists between known mAP and unknown recall (U-Rec). 
Methods focusing on unknown detection often sacrifice known class performance by misclassifying known objects as unknown, while methods maintaining high known mAP tend to have poor unknown recall. The H-Score, measuring the harmonic mean of both metrics, reflects the model's ability to accurately separate known and unknown objects.
As shown in \Cref{tab:main}, our DEUS achieved the best H-Score performance across all tasks. In particular, DEUS showed strong unknown detection capability, achieving U-Recall scores of 65.1, 66.2, and 69.0 for Tasks 1–3 on M-OWODB, which clearly outperformed other methods. These large improvements in unknown detection were achieved while maintaining competitive known mAP performance, resulting in significantly improved H-Scores that demonstrated our EUS method's superior ability to distinguish known from unknown objects.
Additionally, while other methods showed degraded known mAP as tasks progressed, DEUS maintained stable performance throughout the incremental learning process, showing better resistance to catastrophic forgetting through our EKD loss. From Task 3 onward, DEUS achieved the best performance in all known mAP metrics, showing that our approach successfully reduced cross-influence between previously learned and newly learned classes.

As shown in \Cref{tab:RS-OWOD}, DEUS outperformed OrthogonalDet on RS-OWODB, nearly doubling the H-Score in Task 1 (62.5 vs. 34.8) while maintaining higher known mAP across all tasks, demonstrating that both EUS and EKD generalize effectively beyond natural image domains.
\begin{table}[t]
\caption{Experimental results on RS-OWODB. The best performance is highlighted in bold.}
\resizebox{\columnwidth}{!}{%
\setlength{\tabcolsep}{2pt}
\renewcommand{\arraystretch}{1.4}
\begin{tabular}{l|cgy|cgy|cgy|c}
\specialrule{2pt}{0pt}{0pt}
Task IDs & \multicolumn{3}{c|}{Task 1} & \multicolumn{3}{c|}{Task 2} & \multicolumn{3}{c|}{Task 3} & Task 4 \\
\specialrule{1pt}{0pt}{0pt}
Method & \renewcommand{\arraystretch}{1.0}{\begin{tabular}[c]{@{}c@{}}Known\\mAP\end{tabular}} & U-Rec & H-Score & \renewcommand{\arraystretch}{1.0}{\begin{tabular}[c]{@{}c@{}}Known\\mAP\end{tabular}} & U-Rec & H-Score & \renewcommand{\arraystretch}{1.0}{\begin{tabular}[c]{@{}c@{}}Known\\mAP\end{tabular}} & U-Rec & H-Score & \renewcommand{\arraystretch}{1.0}{\begin{tabular}[c]{@{}c@{}}Known\\mAP\end{tabular}} \\
\hline
OrthogonalDet~\cite{sun2024exploring} & 65.4 & 23.7 & 34.8 & 62.3 & 8.9 & 15.6 & 62.4 & 9.3 & 16.2 & 64.2 \\
\hdashline
\textbf{DEUS (Ours)} & \textbf{67.7} & \textbf{58.1} & \textbf{62.5} & \textbf{65.6} & \textbf{28.1} & \textbf{39.4} & \textbf{66.0} & \textbf{29.7} & \textbf{40.9} & \textbf{68.3} \\
\specialrule{2pt}{0pt}{0pt}
\end{tabular}%
}
\label{tab:RS-OWOD}
\end{table}
\begin{table*}[t]
\caption{Ablation study of DEUS on M-OWODB. EUS and EKD represent ETF-Subspace Unknown Separation and Energy-based Known Distinction loss, respectively. EUS aims to improve the detection performance for unknown objects, while EKD is designed to enhance the performance for known objects.
The best performance is highlighted in bold, with the second-best performance underlined.}
\resizebox{\textwidth}{!}{%
\setlength{\tabcolsep}{6pt} % Default value: 6pt
\renewcommand{\arraystretch}{1.4} % Default value: 1
\begin{tabular}{cc|cgy|cccgy|cccgy|ccc}
\specialrule{2pt}{0pt}{0pt}
\multicolumn{2}{c|}{Task IDs} & \multicolumn{3}{c|}{Task 1} & \multicolumn{5}{c|}{Task 2} & \multicolumn{5}{c|}{Task 3} & \multicolumn{3}{c}{Task 4} \\
\specialrule{1pt}{0pt}{0pt}
EUS & EKD &
\renewcommand{\arraystretch}{1.2}{\begin{tabular}[c]{@{}c@{}}Current\\mAP\end{tabular}} & U-Rec & H-Score &
\renewcommand{\arraystretch}{1.2}{\begin{tabular}[c]{@{}c@{}}Previous\\mAP\end{tabular}} &
\renewcommand{\arraystretch}{1.2}{\begin{tabular}[c]{@{}c@{}}Current\\mAP\end{tabular}} &
\renewcommand{\arraystretch}{1.2}{\begin{tabular}[c]{@{}c@{}}Known\\mAP\end{tabular}} & U-Rec & H-Score &
\renewcommand{\arraystretch}{1.2}{\begin{tabular}[c]{@{}c@{}}Previous\\mAP\end{tabular}} &
\renewcommand{\arraystretch}{1.2}{\begin{tabular}[c]{@{}c@{}}Current\\mAP\end{tabular}} &
\renewcommand{\arraystretch}{1.2}{\begin{tabular}[c]{@{}c@{}}Known\\mAP\end{tabular}} & U-Rec & H-Score &
\renewcommand{\arraystretch}{1.2}{\begin{tabular}[c]{@{}c@{}}Previous\\mAP\end{tabular}} &
\renewcommand{\arraystretch}{1.2}{\begin{tabular}[c]{@{}c@{}}Current\\mAP\end{tabular}} &
\renewcommand{\arraystretch}{1.2}{\begin{tabular}[c]{@{}c@{}}Known\\mAP\end{tabular}}\\
\specialrule{1pt}{0pt}{0pt}
           &            & \underline{66.0} & \underline{36.8} & \underline{47.2} & 58.8          & 45.3          & 52.0          & 29.0          & 37.3 
                        & 52.7          & \underline{43.5} & 49.7          & 30.3          & 37.6          & 49.2          & 31.3          & 44.7          \\
\cdashline{6-18}
           & \checkmark & \underline{66.0} & \underline{36.8} & \underline{47.2} & \underline{59.2} & \textbf{45.9} & \underline{52.6} & 40.0          & 45.4 
                        & \textbf{53.6} & \textbf{43.6} & \textbf{50.3} & \underline{38.9} & 43.9          & \underline{50.3} & \underline{32.7} & \underline{45.9} \\
\hline
\checkmark &            & \textbf{66.2} & \textbf{65.1} & \textbf{65.6} & 58.8          & 45.0          & 51.9          & \underline{63.8} & \underline{57.2} 
                        & 52.6          & 42.4          & 49.2          & \textbf{69.0}  & \underline{57.5} & 48.1          & 29.7          & 43.5          \\
\cdashline{6-18}
\checkmark & \checkmark & \textbf{66.2} & \textbf{65.1} & \textbf{65.6} & \textbf{61.0} & \underline{45.7} & \textbf{53.3} & \textbf{66.2} & \textbf{59.0} 
                        & \underline{53.4} & 43.3          & \underline{50.1} & \textbf{69.0} & \textbf{58.0} & \textbf{50.5} & \textbf{32.8} & \textbf{46.0} \\
\specialrule{2pt}{0pt}{0pt}
\end{tabular}%
}
\label{tab:ablation}
\end{table*}

\subsection{Ablation Study}
\Cref{tab:ablation} summarizes the ablation study on M-OWODB. Note that the baseline (without EUS and EKD) differs slightly from the OrthogonalDet results in \Cref{tab:main}, as it incorporates our improved pseudo-labeling process (see Appendix A). EUS significantly enhanced U-Recall but slightly reduced known mAP due to increased unknown detections. EKD consistently improved performance across all tasks regardless of whether EUS was applied. When both were applied, DEUS achieved the highest H-Score on every task.

\begin{figure}[t]
    \centering
    \begin{subfigure}[b]{\columnwidth}
        \centering
        \includegraphics[width=0.85\columnwidth]{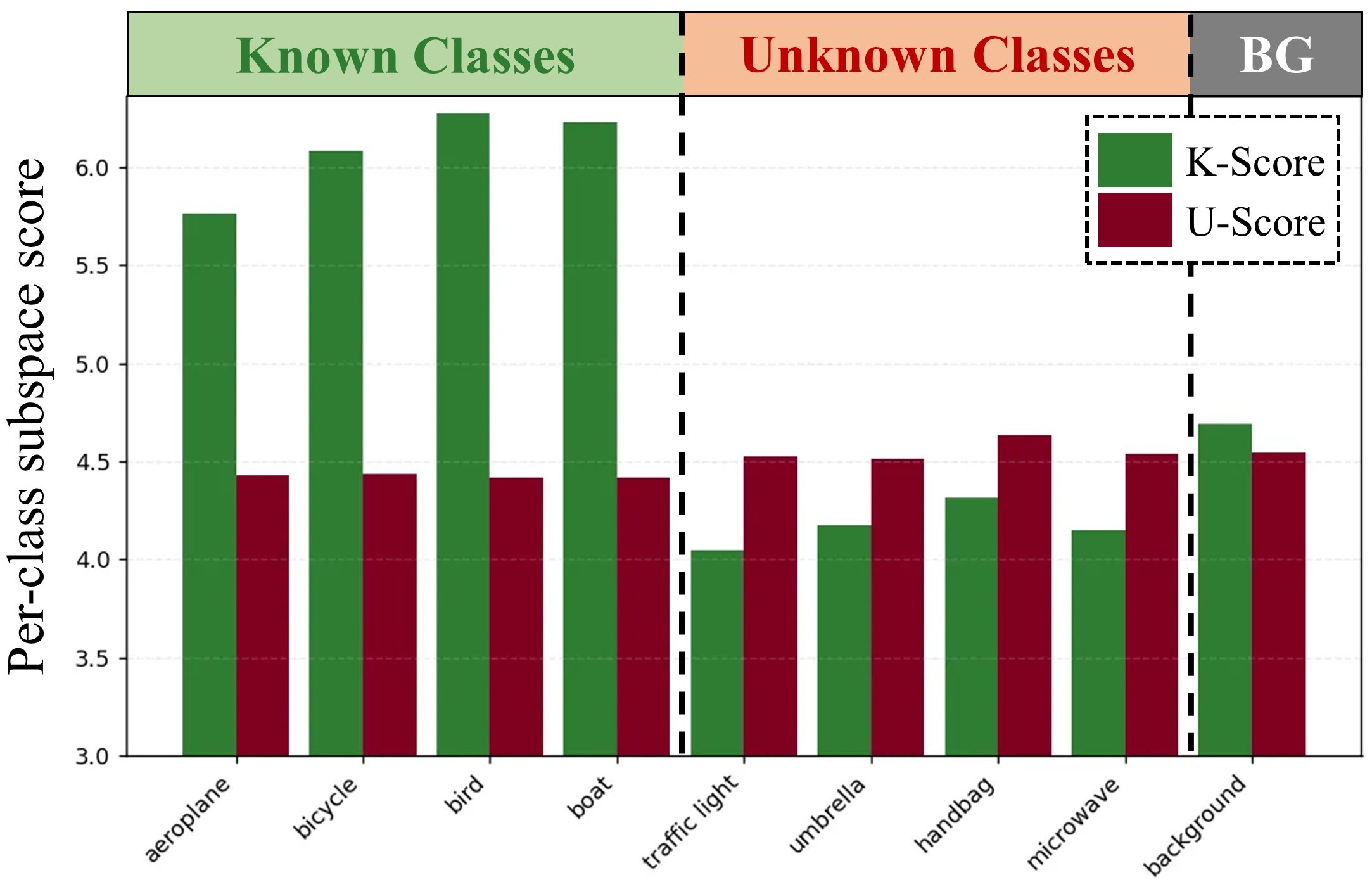}
        \caption{Subspace scores per class.}
        \label{fig:score_a}
    \end{subfigure}
    \vspace{1mm}
    \begin{subfigure}[b]{\columnwidth}
        \centering
        \includegraphics[width=\columnwidth]{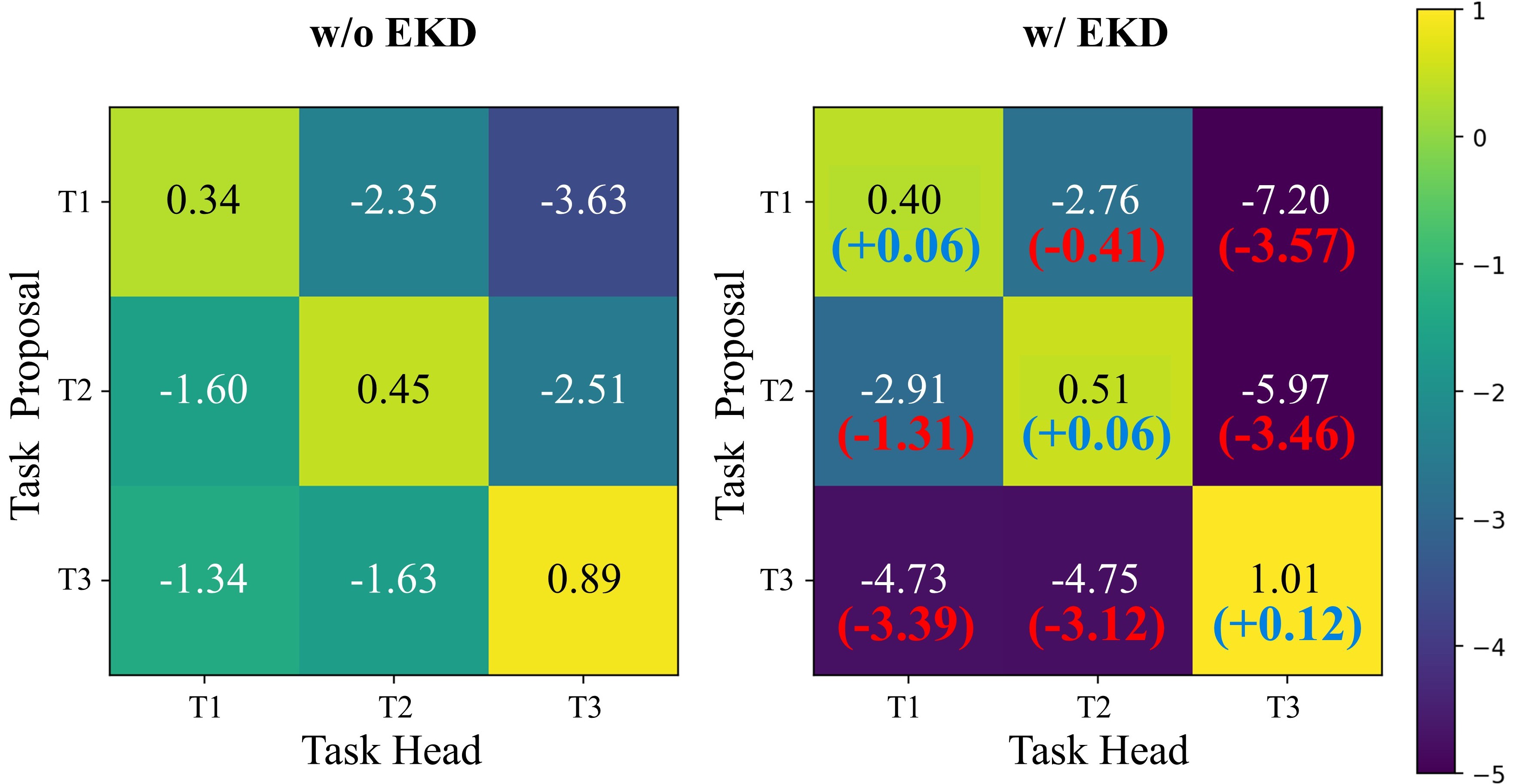}
        \caption{Energy scores by each task head.}
        \label{fig:score_b}
    \end{subfigure}
    \vspace{-8mm}
    \caption{(a) Per-class subspace score comparison between known (green) and unknown (red) for proposals matched with ground-truth. (b) Energy scores of proposals from each task computed by each task head (w/o vs. w/ EKD). Parentheses show changes with EKD.}
    \label{fig:Score}
    \vspace{-4.5mm}
\end{figure}

\subsection{Analysis}
To validate how well EUS separates known and unknown objects, we analyzed the per-class subspace scores for proposals matched to actual ground-truth objects using a model trained on Task 1 classes (known classes 1--20), as shown in Fig.~\ref{fig:score_a}. Proposals matched to known classes consistently show higher known scores than unknown scores, with strong affinity toward the known subspace. In contrast, proposals matched to unseen classes exhibit higher unknown scores, indicating proper alignment with the unknown subspace. Background proposals naturally fall at marginal scores between the two subspaces. This clear separation enables effective distinction between known, unknown, and background objects, as also visualized in the bottom of Fig.~\ref{fig:motiv_space}.

Furthermore, Fig.~\ref{fig:score_b} visualizes the EKD energy score heatmap (Eq.~\ref{head_score}) across task heads. With EKD, diagonal scores (matching task-head pairs) increased while off-diagonal scores were significantly suppressed (e.g., T3-T3: 0.89$\rightarrow$1.01; T1-T3: -3.63$\rightarrow$-7.20), demonstrating that EKD effectively guides proposals towards their appropriate heads. Additionally, DEUS introduces minimal overhead: +1.9\% inference time, +0.5\% FLOPs, and +6.2\% training time over OrthogonalDet (see Appendix D).

Fig.~\ref{fig:qualitative_comparison} compares OrthogonalDet and DEUS on a Task 1 scene where cow is known and giraffe is unknown. OrthogonalDet misclassifies a giraffe as ``horse,'' labels a cow as ``unknown,'' and misses the central unknown animal. DEUS correctly identifies all giraffe instances as ``unknown,'' properly recognizes the cow, and detects the missed unknown.

\begin{figure}[t]
    \centering
    \begin{subfigure}[b]{\columnwidth}
        \centering
        \raisebox{0.45\height}{\rotatebox{90}{\textbf{OrthogonalDet}}}\hspace{1mm}%
        \includegraphics[width=0.88\columnwidth]{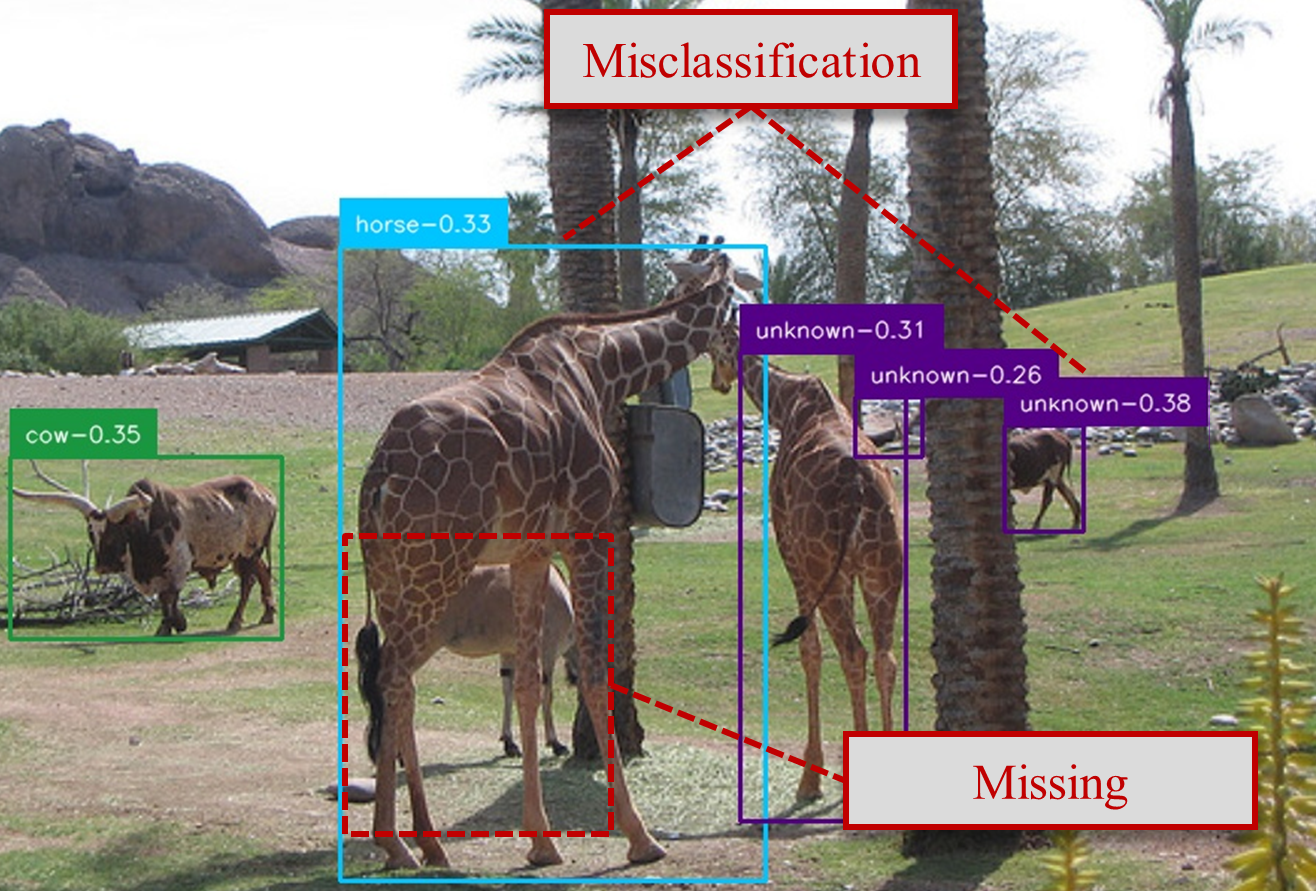}
    \end{subfigure}
    \vspace{1mm}
    \begin{subfigure}[b]{\columnwidth}
        \centering
        \raisebox{0.45\height}{\rotatebox{90}{\textbf{DEUS (Ours)}}}\hspace{1mm}%
        \includegraphics[width=0.88\columnwidth]{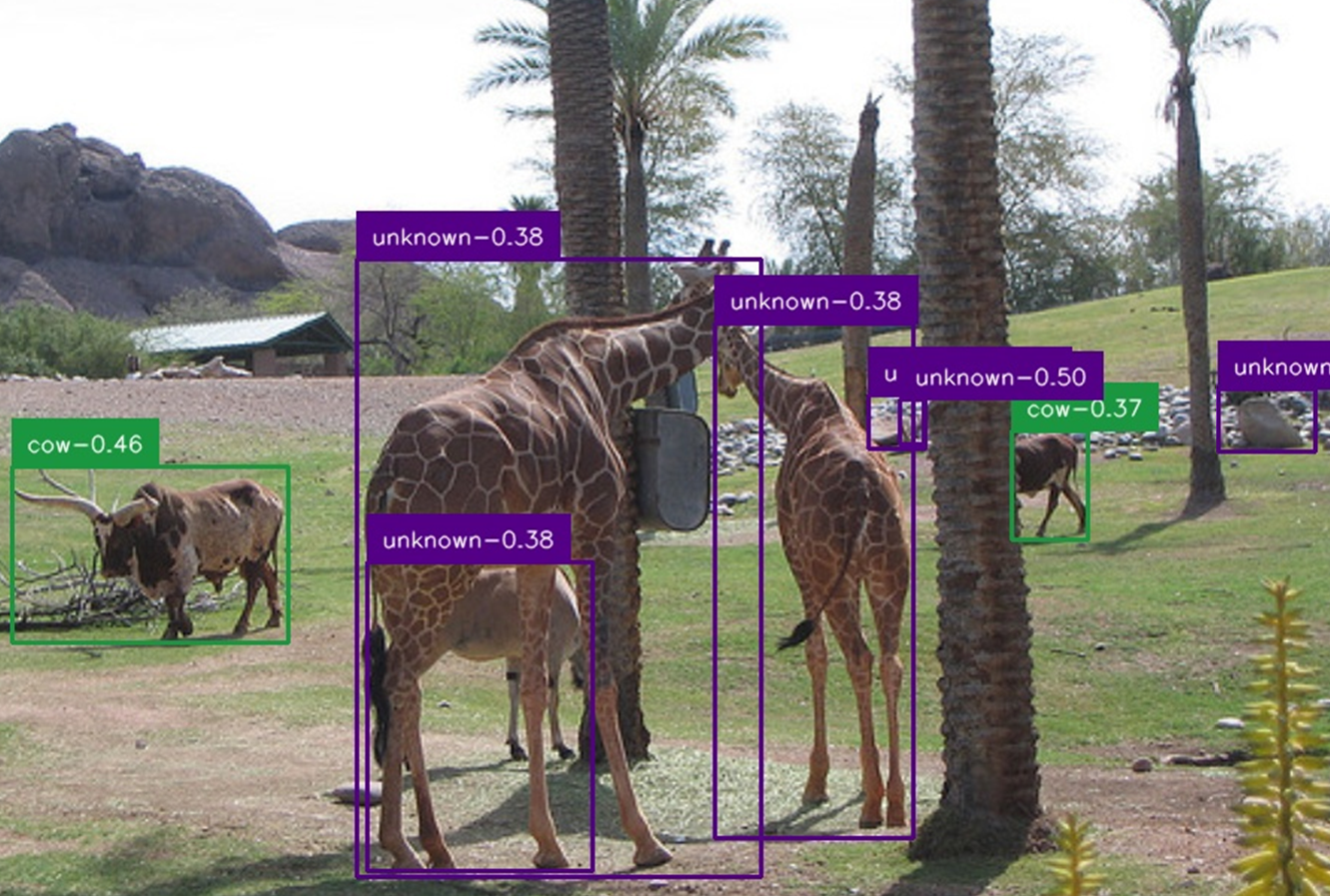}
    \end{subfigure}
    \vspace{-2mm}
    \caption{Qualitative comparison between OrthogonalDet and DEUS. Purple boxes indicate unknown predictions, while other colored boxes indicate known class predictions. The model was trained on Task 1, where cow is known and giraffe is unknown.}
    \label{fig:qualitative_comparison}
    \vspace{-3.5mm}
\end{figure}
\section{Conclusion}
In this paper, we propose DEUS, a novel framework for Open World Object Detection (OWOD). OWOD requires models to learn known classes incrementally while detecting unknown objects, which raises challenges such as limited representation learning of unknown objects and cross-influence between old and new classes during memory replay. DEUS addresses these issues through two modules: ETF-Subspace Unknown Separation (EUS), which captures representations of unknown objects by separating them from known, and Energy-based Known Distinction (EKD), which mitigates cross-influence by focusing on each class set. Nonetheless, semantic overlap between known and unknown objects remains challenging, motivating future work on more refined representation learning.

\clearpage
\section*{Acknowledgments}
This research was conducted with the support of the HANCOM InSpace Co., Ltd. (Hancom-Kyung Hee Artificial Intelligence Research Institute), and supported by the ``Advanced GPU Utilization Support Program'' funded by the Government of the Republic of Korea (Ministry of Science and ICT), and supported by Institute of Information \& communications Technology Planning \& Evaluation (IITP) grant funded by the Korea government (MSIT) (RS-2019-II190079, Artificial Intelligence Graduate School Program (Korea University), and RS-2024-00457882, AI Research Hub Project).
% \clearpage
% --------- References ---------
{
    \small
    \bibliographystyle{ieeenat_fullname}
    \bibliography{main}
}

% WARNING: do not forget to delete the supplementary pages from your submission
% \input{sec/X_suppl}

\end{document}